# Search Using N-gram Technique Based Statistical Analysis for Knowledge Extraction in Case Based Reasoning Systems

M. N. Karthik
3rd Year ECE,
NITS, University of Madras
India
+91 44 4990092
mnk@acm.org

Moshe Davis
3rd Year ECE,
NITS, University of Madras
India
+91 98411 99073
moshedavis@hotmail.com

**ABSTRACT**

Searching techniques for Case Based Reasoning systems involve extensive methods of elimination. In this paper, we look at a new method of arriving at the right solution by performing a series of transformations upon the data.

These involve N-gram based comparison and deduction of the input data with the case data, using Morphemes and Phonemes as the deciding parameters.

A similar technique for eliminating possible errors using a noise removal function is performed. The error tracking and elimination is performed through a statistical analysis of obtained data, where the entire data set is analyzed as sub-categories of various etymological derivatives. A probability analysis for the closest match is then performed, which yields the final expression. This final expression is referred to the Case Base. The output is redirected through an Expert System based on best possible match. The threshold for the match is customizable, and could be set by the Knowledge-Architect.

## INTRODUCTION

The domain of Case Based Reasoning requires comparison of Input Data with those in the Case Base. Most methods employed for this are static and inflexible beyond a certain stage. For data originating from a domain containing highly varied set of elements, as in the arena of computers we are in need of method that is capable of comparing the data through a set of transformations.  Here the method employed is flexible to a very great extent, and the efficiency depends on the conditions set by the Knowledge Engineer. In this method, the data is primarily converted into a form understandable by the CBR system, and is then subjected to a set of transformations.

In this paper, we use the N-gram technique as the basis for the data transformation, extraction, sampling and comparison. A brief note on the N-gram technique is provided at the end of the paper as Appendix - I.

A standard sample Input Data string is denoted by the variable ID and a Case Data string is denoted by the variable CD. The N-gram function of $k^{th}$ degree using Morphemes is denoted by the notation N-gram (k, M, String) and that using Phonemes is denoted by N-gram (k, P, String).

The contents of the ID & CD form the FSM, where each N-gram combination has a corresponding variable denoting the probability of occurrence of that combination. Analysis of these combinations is performed to arrive at the best match.

## PRE-PROCESSING

Let us consider a support scenario wherein the user defines the entire problem in a single string. Consider the following examples –

**ID = THE SYSEM HANGING WHEN DOING INSTALLATION**   (System has been necessarily misspelled to Sysem)

**CD = SOFTWARE CRASHES WHEN I RUN A PROCESS**

The purpose of our search technique is to simplify the ID to such a degree that a suitable comparison and assessment with respect to the CD could be made.

The first step involved in this is to perform an Etymological Derivation on the words, and using the root words of the ID. This helps eliminate morphological changes performed on the data. An assumption made here is that the CD need not necessarily be simplified in anyway.

For the purpose of Etymological Derivation, the system uses a preset lexicon. The lexicon exists as a simple database showing synonym relationships. In this stage, the derived words replace the actual words in the ID.

Hence after the Etymological Derivation, the ID now becomes –

**ID = THE SYSEM *HANG* WHEN DO *INSTALL***

**CD = SOFTWARE CRASHES WHEN I RUN A PROCESS**

The changes are highlighted by the italic, namely HANG and INSTALL from HANGING and INSTALLATION, respectively. Once the pre-processing is performed, the other advanced data transformations can be performed at a much faster rate for comparison with the data in the Case Base.

**STATISTICAL ANALYSIS USING N-GRAM TECHNIQUE**

After the Etymological Derivation, the words obtained are processed through simple filters, for the elimination of common words, prepositions, conjunctions and the like. This could involve mistakes too, and the Knowledge Engineer can make exceptions for words that may not be mistakes, but would be proper nouns. Examples of this would be product names, application names, etc. The parser could automatically treat them so, and ignore them from any further processing in this stage. The other words are then redirected through a simple noise filter to remove the unwanted words.

At the end of this, the data may look like –

**ID = SYSEM *HANG* DO *INSTALL***

**CD = SOFTWARE CRASHES WHEN I RUN A PROCESS**

The next step is to perform a series of N-gram technique based analysis on the ID, and compare them statistically with existing Case Base Data. Here, the N-gram technique is applied on the ID using a context specific set of rules. This would also involve error correction, based on the outcome of the N-gram based analysis.

The following factors are used in the analysis –

1. Phonetic similarity - This refers to the amount of phonetic difference between the word input by the user and the word conjured up by the system.

2. Lexicon similarity - This is the same as the previous one, except that here it corresponds to a linguistic word, rather than a phonetic one.

3. Context sensitive - This checks the context of the word with respect to the other words in the sentence. For instance, a question on a hardware related problem is more probable to contain words corresponding to hardware. Since other words in the system will also contain pointers to the same, short-listing the possibilities becomes easy when the word is looked in a specific domain.

4. Domain sensitive - This is similar to the previous one, except that instead of checking the probability of the word occurring with respect to the words of the sentence, it is checked with the words making up the domain of the problem. If the question were posed to a domain of computer specific expert systems, the application would check it up with all related words in the domain of computers.

Each of the above factors is assigned a weightage, and the mean weightage of the outcome is taken. The following table shows a sample outcome for the word **SYSTEMS** –

| Data | Systems |
|---|---|
| Etymology | System |
| Phoneme constituents | { /s/ /y/ /s/ /t/ /e/ /m/ } |
| Morpheme constituents | { system, s } |
| N-gram on Etymology  From 1 to N-1 (N=5) | 2-gram- {sy, ys, st, te, em } 3-gram- {sys, yst, ste, tem} 4-gram- {syst, yste, stem} 5-gram- {syste, ystem } |
| Pragmatic Knowledge (sample keywords) | Computer, machine, Device, Software, etc. |

A similar analysis for the word **SYSEM** would yield the following –

| Data | Sysem |
|---|---|
| Etymology | Sysem |
| Phoneme constituents | { /s/ /y/ /s/ /e/ /m/ } |
| Morpheme constituents | { sysem } |
| N-gram on Etymology  From 1 to N-1 (N=4) | 2-gram- {sy, ys, se, em } 3-gram- {sys, yse, sem } 4-gram- {syse, ysem } |
| Pragmatic Knowledge (sample keywords) | None Available |

A simple analysis of the above data immediately arrives at the conclusion that the probability of **SYSEM** being **SYSTEM** is very high indeed. If there were more than one set of possible data, the average of the weightages obtained is taken and the word with the highest weightage is substituted in place of the possibly erratic word. The Knowledge Architect could also give a higher weightage to any of the 4 parameters, depending on how most mistakes originate.

At the end of this stage, the ID would now be –

**ID = SYSTEM HANG DO INSTALL**

In the case of the example considered, the parser automatically replaces Sysem with System.

**LEXICON BASED SYNONYM REPLACEMENT**

This stage involves the replacement words obtained from the previous stage with their respective synonyms as defined by the Knowledge Engineer. These would be words corresponding to those in the Case Base. Hence, essentially this stage consists of transforming the data previously processed into keywords that are defined in the Case Base.

The synonyms could be defined based on the context too, either based on the words making up the sentence, or based on the domain. For the above-mentioned examples the lists of synonyms would be domain based and they could be –

| INPUT DATA | SYNONYM IN LEXICON |
|---|---|
| System | Software |
| Hang | Crash |
| Install | Run |

Hence if **n** is the root of the word **m**, which is input from the user, **n** is checked against a list of synonyms.

Let **p** the synonym corresponding to **n**. If **m** does not fall in any exception list, **m** is checked with a synonym set.

Consider a set of words **A {p, q, r}** forming a synonym set, i.e. **p, q, r** all mean the same. The knowledge architect could choose to replace each of these occurrences with yet another synonym, say **s**. The data stored in the Knowledge Base would be represented in terms of s, so identifying the data becomes easy.

So the data would now be -

**ID = SOFTWARE CRASH DO RUN**

**CD = SOFTWARE CRASH WHEN I RUN A PROCESS**

The Input Data is strikingly nearer to the Case Data, which makes for easier comparison.

**NOISE REMOVAL & FINAL COMPARISON**

During the Noise Removal stage, words, which are marked as noise by the Knowledge Engineer, are removed. By default, these would be prepositions, conjunctions and articles. This could also be extended to qualifiers and modifiers like adverbs, adjectives, etc.

The words could also be user defined, and domain specific ones could be included in this category. In an instance, just adding the word *car* as noise to a CBR system for *cars* shot up the efficiency from about 55% to more than 81%.

In this stage, the operations are performed both on the ID and the CD, unlike other operations, where only the ID changes.

Performing a simple linguistic Noise Removal on the example data yields –

**ID = SOFTWARE CRASH RUN**

**CD = SOFTWARE CRASH RUN PROCESS**

To establish the final data correlation, N-gram technique is again applied to the string constituents of the data.

Based on this, weightages are assigned and the solution is that case with the highest weightage.

Since the data is now in an easier form to enable better comparison, the solution becomes a lot more accurate.

In case of the example considered, we see that the ID is a subset of CD, hence the solution for CD could be assigned to the problem of ID.

Upon closer inspection of the ID and CD, we notice that the data is now in a much simplified form. This would also enable us to convert all data in the Case Base to an understandable format using the same transforms so that further cases would automatically fit in. But doing this may limit the system performance if the Knowledge Engineer wants to assign more than one synonym and hence compare the ID with more than one CD.

**CONCLUSION**

We see how this technique helps speed up the searching methodologies employed in Case Based Reasoning Systems. This technique can be further enhanced to incorporate modified data into the Knowledge Base as well as the Case Base.

The lexical database could also be made to accommodate frequent errors so that the processing time is saved.

A good implementation of this system would yield atleast a 30% improvement of searches in Case Based Reasoning systems. In congruence with other advanced search algorithms and Natural Language Processing techniques, this technique could yield significant performance improvements.

**APPENDIX-I (N-GRAM TECHNIQUE)**

The N-Gram technique involves splitting the words into N parts heuristically, and checking each of the combinations with a similar combination on words from a lexicon.

For example, consider the user input string & case string -

**ID = CONTRACTED**

**CD = CONTACT**

Applying a 3-Gram technique, the words are split into groups of three letters as given below with the number of times that particular group occurring in the word.

In the above case, that would yield -

**ID = {(CON,1), (ONT,1), (NTR,1), (TRA,1), (RAC,1), (ACT,1), (CTE,1),(TED,1)}**

CD = {(CON,1), (ONT,1), (NTA,1), (TAC,1), (ACT,1)}

For each of this, a weightage is calculated, which is know as the Score. The Score is given by -

**Score = 100 * Sum ((minimum counts of common grams)/(maximum number of grams))**

Using this in the above-mentioned strings, we see three sets are common, which are -

**(CON, 1), (ONT, 1), (ACT, 1)**

The score for this could be calculated as -

**Score = 100 * (Sum (1 + 1 + 1)/10)**

The maximum number is calculated through the number of unique sets obtained. In this case that would be Score = 30. This score would determine the similarity of the strings, essentially higher the Score, higher the similarity.

The same procedure is used in comparing the words in the dictionaries, both phonetic and linguistic or words lexicons.

Here we consider the words that were discussed earlier -

**ID = SYSEM**

Applying a 2-Gram technique to this would split it into groups of 2 as

**N-gram (2, M, SYSEM) = {(SY,1), (YS,1), (SE,1), (EM,1)}**

This is then compared with the groups of the word SYSTEM. Comparing this with a word lexicon (not phonetic) gives -

**N-gram (2, M, SYSTEM) = {(SY,1), (YS,1), (ST,1), (TE,1), (EM,1)}**

Hence, the Score of this would be

**Score = 100 *((1 + 1 + 1)/6) = 50**

This is a very high score for a 5-7 letter word comparison. Hence, the probability for Sysem to be System is very high in the word lexicon comparison. A similar result would be obtained in the phonetic lexicon comparison. A ratio could also be calculated, which would be a function of the number of letters in the words in both cases and the score.

An example of this would be taking the mean of the number of letters in both the words and finding the ratio of that with the score. The ratio could be standardized on a 10 scale or a 100 (percentile) scale. For the example considered above, a standardized value on a percentile scale would give 91%. This is a very high percentile, and hence that it is highly probable that the word is indeed **SYSTEM**, and not **SYSEM**.

Using this comparison technique for the above two phrases, the system will pick the CD for the given ID based on the percentile obtained in both word and phonetic lexicons. The Knowledge Engineer could also specify the percentile required or a threshold percentile for a word to even qualify. So based on the threshold value that is set and comparing this value with that obtained, the nearness could be calculated. A higher nearness would mean a higher probability of the user input word being the compared word in the lexicon.

**ACKNOWLEDGMENTS**
We thank the Cybernet Software Systems, Inc. and Slashsupport, Inc. for resources provided and the support rendered.